\documentclass[lettersize,journal]{IEEEtran}
\usepackage{amsmath,amsfonts}
\usepackage{algorithmic}
\usepackage{algorithm}
\usepackage{array}
\usepackage[caption=false,font=normalsize,labelfont=sf,textfont=sf]{subfig}
\usepackage{textcomp}
\usepackage{stfloats}
\usepackage{url}
\usepackage{verbatim}
\usepackage{graphicx}
\usepackage{cite}

\begin{document}

\title{VFM$^{4}$SDG: Unveiling the Power of VFMs for \\ Single-Domain Generalized Object Detection}

\author{
Yupeng Zhang, 
Ruize Han, 
Ningnan Guo, 
Wei Feng~\IEEEmembership{Member,~IEEE,}
Song Wang~\IEEEmembership{Member,~IEEE,}
Liang~Wan~\IEEEmembership{Member,~IEEE}
\thanks{Y. Zhang, N Guo, W. Feng and L. Wan are with the College of Intelligence and Computing, Tianjin University, Tianjin 300350, China. Email: zhangyupeng@tju.edu.cn, guo\_nn@tju.edu.cn, wfeng@tju.edu.cn, lwan@tju.edu.cn}
\thanks{R. Han, and S. Wang are with the Faculty of Computer Science and Artificial Intelligence, Shenzhen University of Advanced Technology, Shenzhen 518107, Guangdong, China. Email: han\_ruize@tju.edu.cn, wangsong@suat-sz.edu.cn}
}

\markboth{IEEE TRANSACTIONS ON IMAGE PROCESSING, IN SUBMISSION} %
{Shell \MakeLowercase{\textit{et al.}}: A Sample Article Using IEEEtran.cls for IEEE Journals}


\maketitle

\begin{abstract}
Real-world weather, illumination, and imaging variations often induce severe domain shifts, degrading single-source detectors in unseen environments.
Existing single-domain generalized object detection (SDGOD) methods mainly rely on data augmentation or domain-invariant learning, while largely overlooking how domain shift disrupts detector prediction stability.
Through analytical experiments, we find that performance degradation is mainly dominated by increasing missed detections. Further analysis shows that this phenomenon stems from reduced cross-domain stability in DETR-style detectors: domain shift disrupts encoder-side object-background and inter-instance relations, and further weakens the semantic-spatial binding between decoder queries and real objects. Motivated by this, we find that vision foundation models (VFMs) still preserve stable relational structures and object responses under severe shifts, making them suitable cross-domain stability priors to compensate for detector degradation.
To this end, we propose VFM$^{4}$SDG, a dual-prior learning framework for SDGOD, which introduces a frozen VFM into encoder representation learning and decoder query modeling. Specifically, we propose Cross-domain Stable Relational Prior Distillation to distill stable object-background and inter-instance relations from the VFM into the encoder, compensating for relational degradation. Meanwhile, we propose Semantic-Contextual Prior-based Query Enhancement, which injects category semantic prototypes and global object context into queries before they enter the decoder layer, enhancing semantic-spatial query-object binding stability. 
Extensive experiments show that VFM$^{4}$SDG significantly outperforms existing advanced methods on standard SDGOD benchmarks and two mainstream DETR-based detection frameworks, demonstrating its effectiveness, robustness, and generality.
\end{abstract}

\section{Introduction}
Most deep object detectors assume that training and test data share the same distribution. However, in real-world visual perception, changes in weather, illumination, and imaging conditions often cause substantial domain shifts, leading to severe degradation in unseen environments. Among domain-generalized detection settings, Single-Domain Generalized Object Detection (SDGOD)~\cite{wu2022single} is particularly practical and challenging: the model is trained on a single source domain, yet must generalize to unseen target domains without target-domain supervision.
Existing SDGOD methods mainly improve robustness through data augmentation, feature disentanglement, or their combination. However, their gains remain limited under severe domain shifts: augmentation with manually designed or predefined style variations can hardly cover complex real-world shifts, while static disentanglement of domain-invariant and domain-specific factors may introduce information loss or additional representation bias. \textbf{Overall, existing methods largely follow the domain generalization paradigm from classification, focusing on appearance diversification or domain-invariant representation learning, but rarely analyze how domain shift destabilizes the detector's internal prediction process.}

\begin{figure}[t!]
	\centering
	\includegraphics[width=1.0\linewidth]{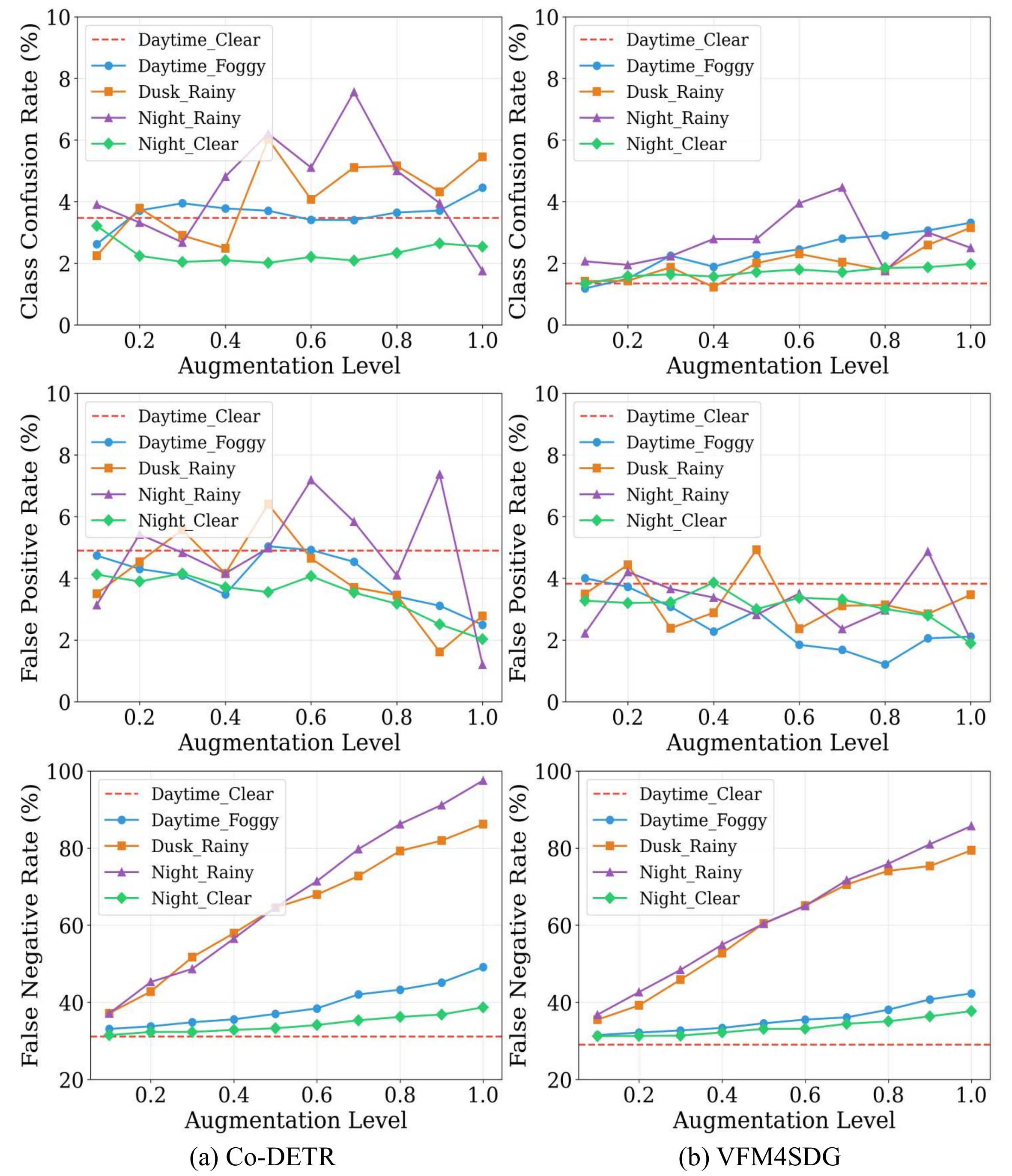}
	\vspace{-23pt}
    \caption{Evolution of error types under increasing synthetic domain shifts, based on 2000 randomly sampled test images. The augmentation level controls the strength of weather/style perturbations and serves as a proxy for domain-shift severity. (a) For Co-DETR, false negatives increase monotonically and dominate the degradation, while class confusion and false positives remain low or even decrease. (b) VFM$^{4}$SDG substantially suppresses the growth of false negatives, indicating that it alleviates miss-dominated degradation and mitigates target-to-background collapse.} 
	\label{fig:moti}
	\vspace{-20pt}
\end{figure}

Recent works such as DG-DETR~\cite{hwang2025dg} and SA-DETR~\cite{han2025style} show that Transformer-based detectors~\cite{carion2020end} hold great promise for SDGOD. By explicitly modeling global context, instance interactions, and query-driven prediction through attention, they also provide a suitable framework for analyzing how domain shift affects detector stability.
Building on this, we further analyze how domain shift affects DETR-style detectors. Specifically, we take daytime-clear images as the source scenario and evaluate progressively stronger shifts, including fog, rain, night, and night-rain. As shown in Fig.~\ref{fig:moti}, the performance degradation is mainly dominated by a continuous increase in false negatives, while category confusion, \textit{e.g.}, misclassifying a car as a truck, remains low and stable, and false positives are also generally low and even decrease in most cases. This indicates that the main degradation in SDGOD is not that real objects are misclassified as other foreground categories, but that their foreground activation gradually weakens and eventually manifests as dominant background / no-object responses. We term this phenomenon \emph{object-to-background collapse}.

\begin{figure}[t!]
	\centering
	\includegraphics[width=1.0\linewidth]{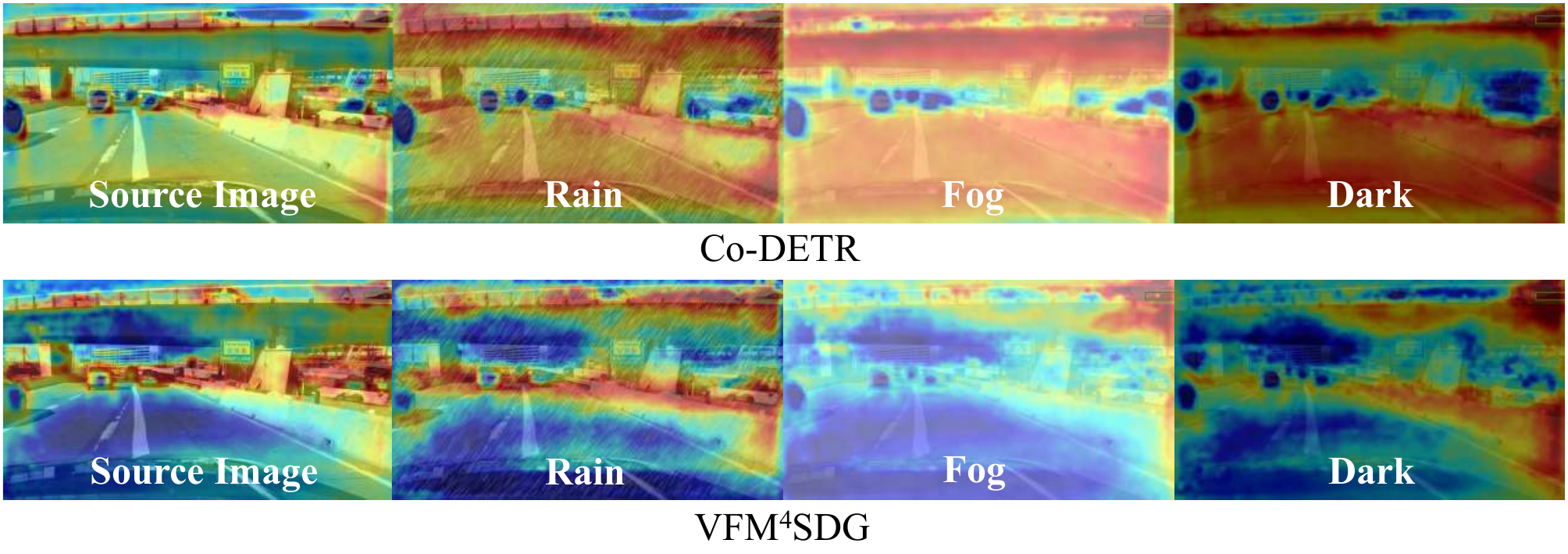}
	\vspace{-23pt}
	\caption{Object-centric correlation visualization in the encoding stage. For each image, one target instance is selected as the anchor, and its correlation with all image tokens is computed based on encoder features. From left to right are the source image and its rainy, foggy, and dark counterparts. Under different domain-shift conditions, Co-DETR produces more diffuse responses over broader regions, indicating weakened object-background discrimination and relational stability. In contrast, VFM$^{4}$SDG preserves more concentrated target-related responses across domains, demonstrating stronger relational stability and domain robustness.} 
	\label{fig:fail}
	\vspace{-20pt}
\end{figure}

From the perspective of DETR-style detection, object-to-background collapse is not merely caused by appearance degradation, but is closely tied to weakened foreground activation in query-driven prediction. Unlike conventional detectors that rely on dense candidates, DETR-style detectors use a limited set of object queries to retrieve semantic and spatial cues from encoder features and establish one-to-one semantic-spatial correspondences with objects. Thus, stable detection depends not only on clear local appearance, but also on stable object-background and inter-instance relations in encoder features.
As shown in Fig.~\ref{fig:fail}, encoder features maintain concentrated object responses in the source domain. However, under shifted conditions such as fog, rain, and darkness, object-related responses spread to background regions and object-background boundaries become blurred, indicating disrupted foreground separability and inter-instance relations. Such upstream relational degradation weakens the semantic-spatial binding between decoder queries and objects, making queries more prone to low foreground confidence or dominant no-object responses, and eventually leading to missed detections. Therefore, the core challenge of SDGOD is not only to learn domain-invariant instance semantics, but also to preserve encoder-side relational stability and mitigate its adverse effect on decoder-side query-object binding.

\begin{figure}[t!]
	\centering
	\includegraphics[width=1.0\linewidth]{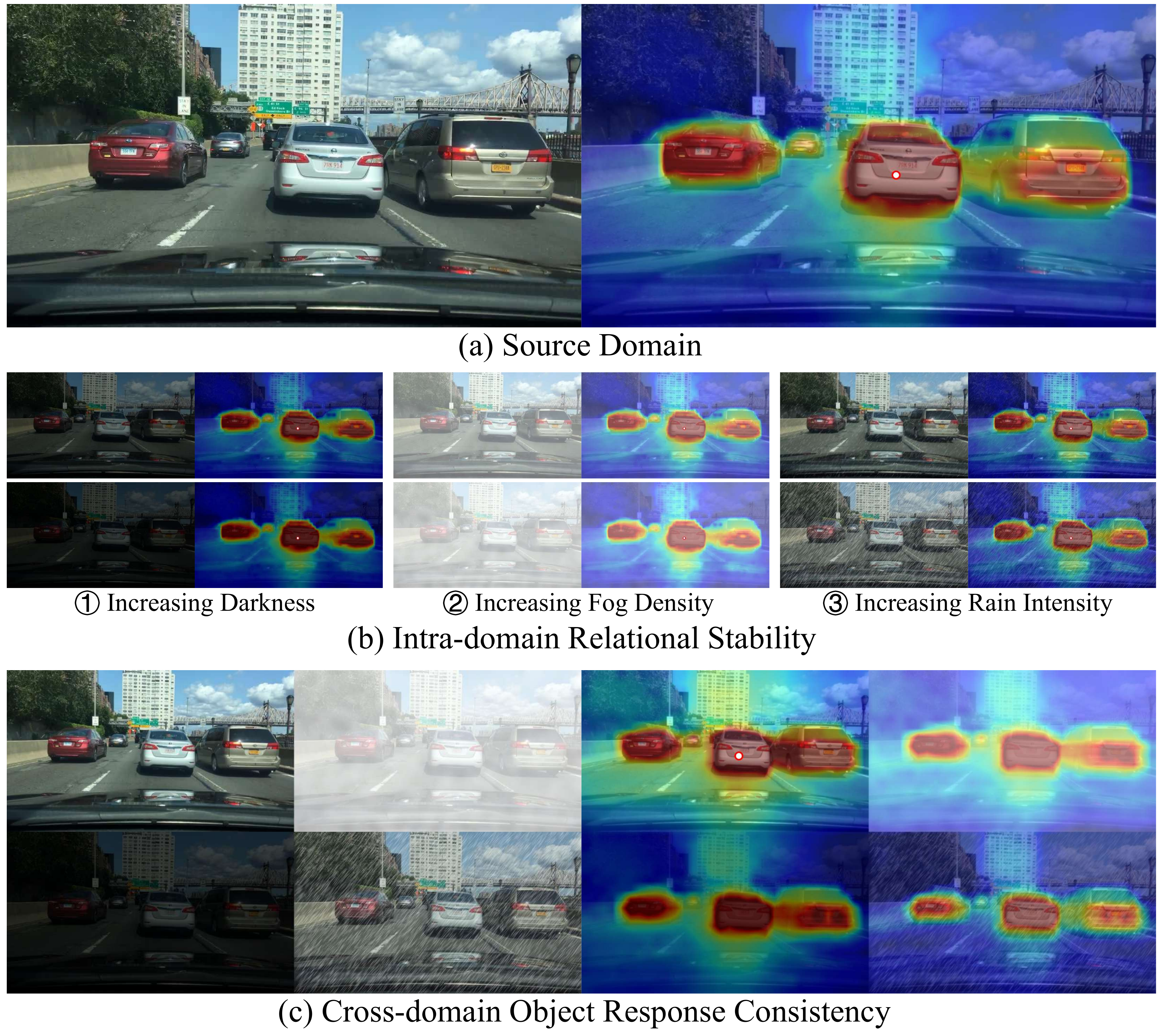}
	\vspace{-23pt}
	\caption{Visualization of DINOv3 stability priors under domain shifts. Given an anchor point on a target object, we compute its feature similarity to all image patches and overlay the response map on the image.
    (a) In the source domain, DINOv3 highlights the anchor object and semantically related instances while preserving clear object-background separation.
    (b) Under increasing darkness, fog density, and rain intensity, DINOv3 maintains intra-domain relational stability by consistently activating related object regions.
    (c) Across shifted domains, DINOv3 exhibits cross-domain object response consistency, producing coherent and salient responses on corresponding objects despite large appearance changes.} 
	\label{fig:dinov3}
	\vspace{-20pt}
\end{figure}

Accordingly, alleviating object-to-background collapse requires a prior that preserves stable relational structures and consistent object responses under appearance variations, thereby compensating for encoder-side relational degradation and providing reliable semantic and spatial cues for decoder queries. 
Motivated by this need, we examine the representation behavior of vision foundation models (VFMs) under severe domain shifts, and find that DINOv3 exhibits cross-domain stability properties highly complementary to the above degradation mechanism. As shown in Fig.~\ref{fig:dinov3}, even under unseen conditions such as fog, rain, and darkness, DINOv3 produces salient responses on real object regions and preserves clear object-background and inter-instance relational structures. 
Specifically, DINOv3 exhibits \textbf{Intra-domain Relational Stability} (Fig.~\ref{fig:dinov3}(b)): objects of the same or semantically related categories maintain stable neighborhoods in feature space across styles, with clear object-background boundaries. This provides a basis for compensating encoder-side relational degradation.
Meanwhile, DINOv3 shows \textbf{Cross-domain Object Response Consistency} (Fig.~\ref{fig:dinov3}(c)): semantically similar object regions still receive spatially consistent and salient responses under different environmental conditions. This provides stable object-level semantic cues and spatial context for decoder queries, helping mitigate query-object binding instability.
Therefore, DINOv3 is not merely used as a stronger semantic teacher; instead, it serves as a degradation-driven transferable cross-domain stability prior that compensates for the relational stability lost by DETR-style detectors in unseen domains and supports more stable query-object binding.

Based on these insights, we propose VFM$^{4}$SDG, a dual-prior guided learning framework for SDGOD, which introduces a frozen VFM as a transferable cross-domain stability prior into encoder representation learning and decoder query modeling.
Specifically, to address encoder-side relational degradation caused by domain shift, we propose \textbf{Cross-domain Stable Relational Prior Distillation (CSRPD)}, which distills stable object-background and inter-instance relations from the VFM into the detector encoder to enhance multi-scale relational modeling. To further mitigate the adverse effect of upstream representation degradation on decoder-side query-object binding, we propose \textbf{Semantic-Contextual Prior-based Query Enhancement (SCPQE)}, which injects VFM-derived category semantic prototypes and global object context into queries before they enter the decoder layer, providing more stable semantic identities and spatial cues. Together, CSRPD and SCPQE compensate for encoder-side relational degradation and enhance decoder-side semantic-spatial query-object binding, effectively alleviating object-to-background collapse under domain shift.

Our contributions are summarized as follows:
\begin{itemize}
\item We reveal a false-negative-dominated degradation pattern in SDGOD, termed \emph{object-to-background collapse}, where domain shift mainly weakens foreground activation rather than causing foreground category confusion. We further show that this stems from disrupted encoder-side object-background and inter-instance relations, which weakens decoder-side query-object binding.
\item We investigate VFMs as cross-domain stability priors for SDGOD and find that DINOv3 preserves two complementary properties under severe shifts: intra-domain relational stability and cross-domain object response consistency, which compensate for the stable structures lost by DETR-style detectors.
\item We propose VFM$^{4}$SDG, a dual-prior framework with Cross-domain Stable Relational Prior Distillation (CSRPD) and Semantic-Contextual Prior-based Query Enhancement (SCPQE). They respectively compensate for encoder-side relational degradation and enhance decoder-side query-object binding, achieving state-of-the-art performance on standard SDGOD benchmarks.
\end{itemize}

\section{Related Work}
\textbf{Single-Domain Generalized Object Detection (SDGOD)} aims to improve detector performance in unseen scenarios using data from only a single source domain.
Existing approaches typically enhance model robustness in unseen domains through carefully designed strategies, which can be grouped into three categories:
(1) Data augmentation. These methods perturb images or features to simulate unseen scenarios and expand the training distribution. CLIP the Gap~\cite{vidit2023clip} leverages CLIP~\cite{radford2021learning} with tailored text prompts to synthesize potential styles and enrich backbone semantic–style representations. DivAlign~\cite{danish2024improving} improves cross-domain robustness through a two-stage `domain diversification + detection alignment' pipeline. NP~\cite{fan2023towards} enhances domains by perturbing shallow feature statistics, while SRCD~\cite{rao2024srcd} employs lightweight self-augmentation and models inter-instance semantic relations to preserve intrinsic structure. 
(2) Feature disentanglement. These methods decouple content and domain-specific factors via tailored architectures or loss functions to suppress domain-sensitive cues. SDGOD~\cite{wu2022single} learns domain-invariant representations through cyclic disentanglement–based self-distillation without labels. DG-DETR~\cite{hwang2025dg} removes domain-induced bias in object queries via orthogonal projection in a style space and separates domain-invariant/variant components through wavelet decomposition. G-NAS~\cite{wu2024g} introduces G-loss to regularize Neural Architecture Search (NAS), preventing overfitting to easily separable features and preserving generalizable ones.
(3) Hybrid strategies. UFR~\cite{liu2024unbiased} disentangles foreground and background and incorporates causal prototype learning to mitigate confounding factors. SA-DETR~\cite{han2025style} projects unseen-domain styles back into the training domain and adopts object-aware contrastive learning to enhance instance-level domain invariance for DETR~\cite{carion2020end}.

Overall, existing methods are still largely inherited from generalization research in image classification, placing greater emphasis on representation invariance while paying less attention to how domain shift acts on detector mechanisms, perturbs internal representations, and degrades prediction stability. In contrast, this work is the first to revisit SDGOD from the perspective of cross-domain detector stability and to introduce VFMs as transferable cross-domain stability priors for detector learning, thereby achieving stronger SDGOD.

\textbf{VFMs and their applications in object detection.} 
In recent years, VFMs can be broadly categorized into supervised models~\cite{touvron2022deit,radford2021learning,li2022grounded,kirillov2023segment,ravi2024sam,carion2025sam} and self-supervised models~\cite{chen2021empirical,caron2021emerging,venkataramanan2023imagenet,yan2023self,he2022masked,bao2021beit,wang2023image,chen2024deconstructing,zhou2021ibot,oquab2023dinov2,simeoni2025dinov3}. Supervised models rely on large-scale annotated data and exhibit strong task transferability, but incur extremely high training and annotation costs that limit scalability. In contrast, self-supervised models eliminate manual annotation and leverage larger datasets to learn generic visual representations via contrastive learning, masked image modeling, or their combination, offering superior scalability. Nevertheless, their downstream performance typically still depends on task-specific fine-tuning.
In object detection, the pre-training–then–fine-tuning paradigm remains the dominant way to exploit VFMs. While it accelerates convergence and improves performance, it usually requires strict architectural alignment between the detector backbone and the VFMs, leading to nontrivial structural adaptations (\textit{e.g.}, ViT-Adapter~\cite{chen2022vision}). Moreover, fine-tuning large-scale VFMs is often impractical under limited computational budgets. To mitigate this issue, several works explore frozen VFMs for detection, but this often requires substantial modifications to detector architectures and training protocols. Frozen-DETR~\cite{fu2024frozen} instead employs CLIP (ViT-L/14-336) as a feature enhancer rather than a backbone, injecting high-level visual semantics into the detector and partially alleviating architectural incompatibility.

Inspired by these efforts, rather than replacing the detector backbone or encoder with a VFM, we treat the VFM as a transferable cross-domain stability prior. \textbf{Direct replacement often requires complex architectural adaptation and multi-scale feature alignment, increasing design and training cost while risking that performance gains are attributed to stronger model capacity or pretrained representations, thereby obscuring how domain shift affects detector mechanisms.} Instead, we preserve the detector's main architecture as much as possible while exploiting the cross-domain stable information encoded in VFMs to guide the detector toward more robust representations and prediction mechanisms under domain shift, offering a new perspective for SDGOD.

\section{The Proposed Method}

\subsection{Preliminaries}
\textbf{Problem Setting.}
Let $D_s$ denote the single source domain containing $N^s$ labeled training images
$\{(x_i^s, y_i^s)\}_{i=1}^{N^s}$, where $x_i^s \in \mathbb{R}^{H \times W \times C}$ is an image and
$y_i^s = \{(b_{ij}^s, k_{ij}^s)\}_{j=1}^{M_i}$ denotes the set of object annotations in image $x_i^s$.
Here, $b_{ij}^s \in \mathbb{R}^4$ is the bounding box of the $j$-th object, $k_{ij}^s \in \{1, \dots, K\}$ is its class label, $M_i$ is the number of objects in image $x_i^s$, and $K$ is the total number of categories.
$H$, $W$, and $C$ denote the image height, width, and number of channels, respectively.

Let $\{D_t\}_{t=1}^{T}$ denote a set of $T$ unseen target domains.
The goal of SDGOD is to train an object detector using only the labeled data from $D_s$ such that it generalizes well to test samples from all target domains $\{D_t\}_{t=1}^{T}$.
We assume that the source and target domains share the same label space.

\begin{figure*}[t!]
	\centering
	\includegraphics[width=0.98\linewidth]{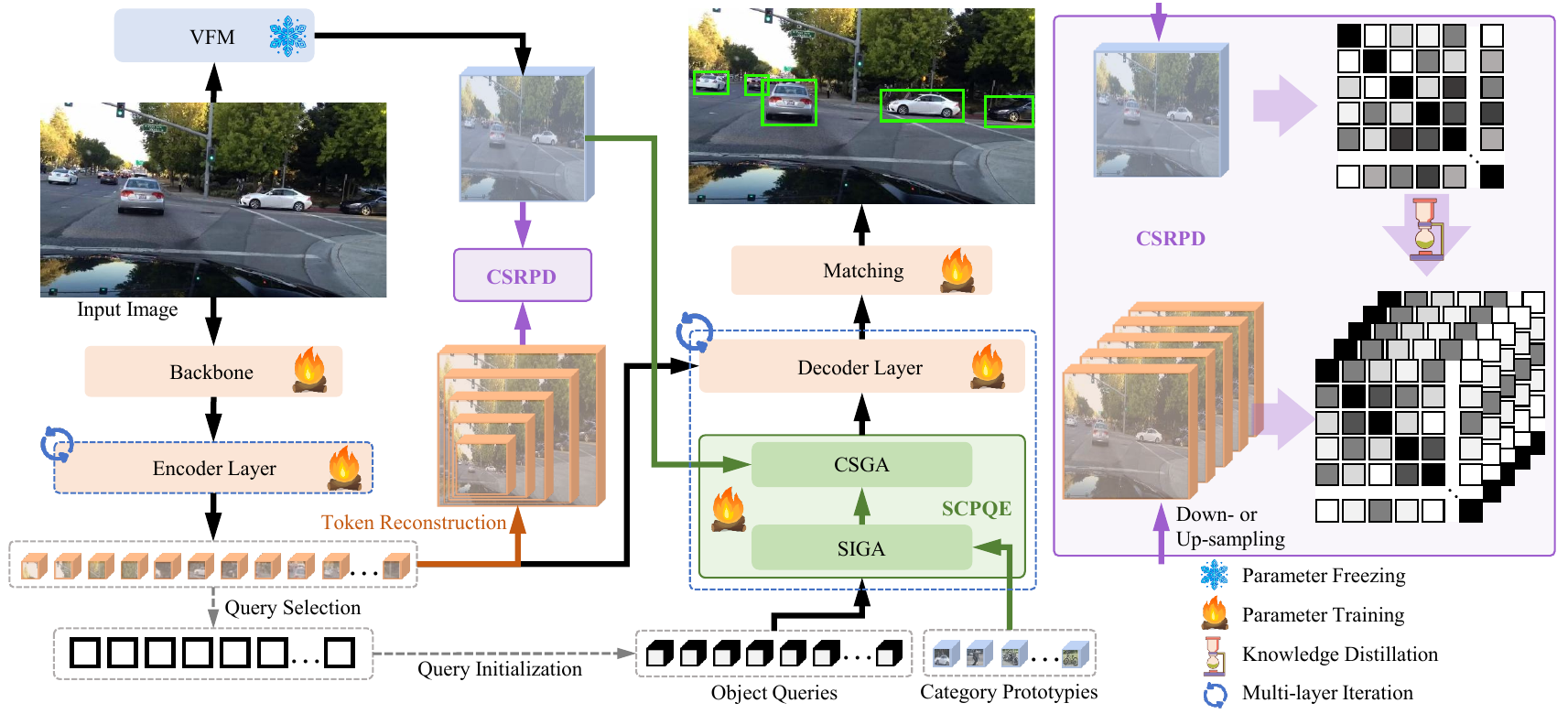}
	\vspace{-10pt}
	\caption{Overall framework of VFM$^{4}$SDG. Built upon a DETR-based detector, VFM$^{4}$SDG leverages a frozen VFM as a cross-domain structural visual prior for SDGOD. At the encoding stage, Cross-domain Stable Relational Prior Distillation (\textbf{CSRPD}) transfers cross-domain stable inter-instance relational structures from VFM to the encoder, yielding a representation space with enhanced relational stability under domain shift. At the decoding stage, Semantic-Contextual Prior-based Query Enhancement (\textbf{SCPQE}) injects VFM-derived semantic and positional priors into decoder queries via cross-attention, enabling stable semantic and spatial query alignment in unseen domains.}  
	\label{fig:overview}
	\vspace{-15pt}
\end{figure*}

\subsection{Overview of the VFM$^{4}$SDG}
As shown in Fig.~\ref{fig:overview}, VFM$^{4}$SDG is built upon a DETR-based detector and treats frozen VFMs as transferable cross-domain stability priors, improving SDGOD from two complementary perspectives: representation learning and query modeling. To this end, we design two tightly coupled modules.
First, in the encoding stage, to alleviate the degradation of object-background and inter-instance relation stability under domain shift, we propose \textbf{Cross-domain Stable Relational Prior Distillation (CSRPD)}. This module constructs token-wise relation matrices from the cross-domain stable relational information encoded by VFMs and distills them into the detector encoder, thereby encouraging relational geometry consistent with the VFM and yielding more stable object-background discrimination and inter-instance relation representations across multi-scale features.
Second, to mitigate the degradation of query semantic-spatial stability under domain shift, we propose \textbf{Semantic-Contextual Prior-based Query Enhancement (SCPQE)}. Specifically, category prototypes are precomputed from source-domain VFM object features and, before queries enter the decoder layer, injected into query representations via cross-attention to establish cross-domain consistent semantic identities. The queries then interact with global VFM features to incorporate more stable semantic and spatial context, thereby improving object recognition and localization in unseen domains.

By introducing stable relational priors in the encoding stage and semantic-contextual priors in the decoding stage, VFM$^{4}$SDG jointly improves detector robustness from both aspects and effectively alleviates severe domain-shift degradation without relying on additional domain augmentation strategies. Below, we take DINOv3 as an example for illustration.
\subsection{Cross-domain Stable Relational Prior Distillation (CSRPD)}
At the encoding stage, to alleviate the degradation of object-background and inter-instance relational representation stability under domain shift, we propose Cross-domain Stable Relational Prior Distillation (CSRPD). By transferring the cross-domain stable relational priors encoded by the DINOv3 into the detector's encoder feature space, CSRPD establishes a more domain-robust relational foundation for subsequent decoder modeling.

\medskip
\noindent\textbf{Teacher feature extraction.}
Given an input image $I$, we first extract dense visual features using a frozen DINOv3 model  as the teacher representation:
\begin{equation}
\mathbf{T} = f_{\text{DINOv3}}(I) \in \mathbb{R}^{B \times C_t \times H_t \times W_t},
\end{equation}
where $B$ denotes the batch size, $C$ denotes the feature channel dimension, $(H_t, W_t)$ denotes the spatial resolution of the DINOv3 features, and $\mathbf{T}$ denotes the teacher feature.

Throughout training, DINOv3 remains fixed, serving only as a teacher that provides cross-domain stable relational priors.

\medskip
\noindent\textbf{Reconstruction of encoder feature pyramids.}
The DETR encoder processes multi-scale features in a flattened token sequence.
To enable relational distillation, we reshape the encoder outputs back to their original pyramid structures.
Specifically, for the $l$-th feature level, the corresponding encoder output is rearranged as
\begin{equation}
\mathbf{X}^{(l)} \in \mathbb{R}^{B \times C \times H_l \times W_l},
\end{equation}
where $(H_l, W_l)$ matches the spatial resolution of the corresponding feature map produced by the backbone.

\medskip
\noindent\textbf{Resolution alignment.}
Since student encoder features and DINOv3 features generally differ in spatial resolution, we align each student feature map to the teacher resolution $(H_t, W_t)$.
For a student feature $\mathbf{X}^{(l)}$, we apply adaptive average pooling when its resolution is higher than the teacher's, and bilinear interpolation otherwise:
\begin{equation}
\tilde{\mathbf{X}}^{(l)} = \mathcal{A}(\mathbf{X}^{(l)}, H_t, W_t),
\end{equation}
where $\tilde{\mathbf{X}}^{(l)}$ denotes the student feature obtained by applying the adaptive alignment operation $\mathcal{A}(\cdot)$ to the original student feature $\mathbf{X}^{(l)}$, resulting in a feature map that is spatially aligned with the teacher and used for subsequent relation matrix computation and distillation.

\medskip
\noindent\textbf{Object-background and inter-instance relational modeling.}
Although the relation matrix is constructed at the spatial token level, we argue that it captures object-background and inter-instance relations. In the DINOv3 feature space, spatial tokens encode not only local semantics but also contextual associations with object and background regions. Tokens from the same instance, though not semantically identical, are typically correlated and jointly represent instance-level semantics, while similarities between object and background tokens reflect object-context structural relations. Therefore, pairwise token similarities can implicitly encode both inter-instance structure and object-background context, rather than isolated local patterns. Empirically, such token-level relational structures remain stable across domains, making them an effective surrogate for modeling structural consistency.

Given a feature map $\mathbf{F} \in \mathbb{R}^{B \times C \times H \times W}$, we flatten it into $N = H \times W$ spatial tokens and apply $\ell_2$ normalization along the channel dimension:
\begin{equation}
\mathbf{Z} = \text{Norm}\!\left(\text{Flatten}(\mathbf{F})\right) \in \mathbb{R}^{B \times C \times N},
\end{equation}
where $B$ denotes the batch size, $C$ the channel dimension, $H$ and $W$ the spatial height and width of the feature map, respectively, and $\text{Norm}(\cdot)$ represents channel-wise $\ell_2$ normalization applied to each spatial token.

We then compute the inter-instance relational matrix by measuring pairwise token similarities:
\begin{equation}
\mathbf{S}(\mathbf{F}) = \mathbf{Z}^{\top} \mathbf{Z} \in \mathbb{R}^{B \times N \times N},
\end{equation}
where each entry encodes the cosine similarity between a pair of spatial tokens.

Accordingly, we obtain the teacher relational matrix $\mathbf{S}_t = \mathbf{S}(\mathbf{T})$ from the DINOv3 feature map $\mathbf{T}$, and the student relational matrices $\mathbf{S}_s^{(l)} = \mathbf{S}(\tilde{\mathbf{X}}^{(l)})$ from the resolution-aligned student features at the $l$-th encoder level.
To prevent trivial self-correlations, the diagonal elements of all similarity matrices are masked out during training.

\noindent\textbf{Cross-domain stable relational prior distillation.}
We distill cross-domain stable relational from the frozen DINOv3 model into the detector encoder across multiple feature levels.
Let $\mathcal{L}$ denote the set of selected encoder feature levels.
The proposed CSRPD loss is formulated as
\begin{equation}
\mathcal{L}_{\text{CSRPD}} =
\sum_{l \in \mathcal{L}}
\ell\!\left(\mathbf{S}_s^{(l)}, \mathbf{S}_t\right),
\end{equation}
where $\ell(\cdot)$ represents the Smooth-$\ell_1$ loss.
In our implementation, relational distillation is applied uniformly across all selected feature levels.

CSRPD is applied exclusively during training and is jointly optimized with the standard detection objective:
\begin{equation}
\mathcal{L} = \mathcal{L}_{\text{det}} + \lambda \mathcal{L}_{\text{CSRPD}},
\end{equation}
where $\mathcal{L}_{\text{det}}$ denotes the original detection loss, and $\lambda$ is a weighting factor that controls the strength of distillation.

\medskip
\textbf{Overall}, by enforcing consistency in object-background and inter-instance relations between the student encoder and the frozen DINOv3 teacher, CSRPD encourages the encoder to preserve cross-domain stable relational representations, thereby enhancing robustness to domain shift and providing a reliable foundation for subsequent decoder modeling.

\subsection{Semantic-Contextual Prior-based Query Enhancement (SCPQE)}
At the decoding stage, to address the degraded semantic-spatial stability of DETR queries under domain shift, caused by their source-domain-biased learning, we introduce Semantic-Contextual Prior-based Query Enhancement (SCPQE). By injecting stable semantic and spatial contextual information, SCPQE improves the queries' ability to recognize and localize objects in unseen target domains.

\medskip
\noindent\textbf{Category prototype construction.}
We first construct category-level semantic prototypes from the source domain using the frozen DINOv3 model.
Given a source-domain image $I_s$ and its ground-truth bounding boxes $\{b_i\}$ with category labels $\{y_i\}$, we extract object-level features from DINOv3 for each annotated region.
Specifically, for an object instance $i$ belonging to category $c$, we obtain its feature vector $\mathbf{v}_i^c$ by spatially pooling the corresponding DINOv3 features within the ground-truth box.
For each category $c$, we aggregate all instance features from the source domain and compute the category prototype as
\begin{equation}
\mathbf{p}_c = \frac{1}{|\mathcal{I}_c|} \sum_{i \in \mathcal{I}_c} \mathbf{v}_i^c,
\end{equation}
where $\mathcal{I}_c$ denotes the set of source-domain instances belonging to category $c$.
By aggregating features from diverse instances, the resulting prototype encodes rich intra-class variations while remaining semantically stable across domains.
All category prototypes are computed offline and stored for subsequent query guidance.

\medskip
\noindent\textbf{Semantic Identity–Guided Attention (SIGA).}
Let $\mathbf{Q} \in \mathbb{R}^{N_q \times C_q}$ denote the initial decoder queries.
To enable effective interaction between queries and category-level semantic priors, we project the category prototypes into the query embedding space using a learnable linear projection:
\begin{equation}
\tilde{\mathbf{P}} = \mathbf{W}_p \mathbf{P},
\end{equation}
where $\mathbf{P} = \{\mathbf{p}_c\}$ denotes the set of category prototypes and $\mathbf{W}_p$ is a learnable projection layer implemented as a single linear layer.
We then inject semantic priors into the queries via cross-attention:
\begin{equation}
\mathbf{Q}_s = \text{LN}\!\left(
\mathbf{Q} + \text{Attn}\!\left(
\mathbf{Q}, \tilde{\mathbf{P}}, \tilde{\mathbf{P}}
\right)
\right),
\end{equation}
where $\text{Attn}(\cdot)$ denotes the standard multi-head cross-attention operation and $\text{LN}(\cdot)$ denotes Layer Normalization.
This operation enables each query to attend to semantically meaningful category-level anchors, providing stable semantic guidance under domain shifts.

\medskip
\noindent\textbf{Contextual Spatial Grounding Attention (CSGA).}
To further inject global semantic and spatial context, we leverage dense image-level features extracted by the frozen DINOv3 model.
Specifically, the DINOv3 feature map is flattened into a sequence of spatial tokens and projected into the query embedding space using another learnable linear projection:
\begin{equation}
\tilde{\mathbf{T}} = \mathbf{W}_t \mathbf{T},
\end{equation}
where $\mathbf{T}$ denotes the DINOv3 feature tokens and $\mathbf{W}_t$ is a learnable linear projection.

The semantically enhanced queries $\mathbf{Q}_s$ then attend to the projected DINOv3 tokens through cross-attention:
\begin{equation}
\mathbf{Q}_p = \text{LN}\!\left(
\mathbf{Q}_s + \text{Attn}\!\left(
\mathbf{Q}_s, \tilde{\mathbf{T}}, \tilde{\mathbf{T}}
\right)
\right).
\end{equation}
This step equips the queries with spatially-aware semantic context that remains stable across domains, thereby improving query alignment robustness in unseen scenarios.

\medskip
\noindent\textbf{Integration with the decoder.}
After sequential semantic and positional prior-guided attention, the refined queries are fed into the standard DETR decoder layer.
By anchoring queries with category-level semantic cues and global visual context, SCPQE stabilizes query representations and enhances semantic–spatial alignment under domain shifts, without introducing additional domain-specific supervision.

\subsection{Implementation Details.}
Our method is built upon the Co-DETR detector and employs a pretrained DINOv3 (ViT-L/16) model as the teacher. To ensure a fair comparison, all experiments consistently adopt ResNet-50 as the backbone. We adopt the AdamW optimizer with a base learning rate of \(2\times10^{-4}\), a weight decay of \(1\times10^{-4}\), and apply a learning-rate multiplier of \(\text{lr\_mult}=0.1\) to the backbone. Gradient clipping with an L2 max-norm of \(0.1\) is employed to stabilize training.
The model is trained for 24 epochs, and we use a MultiStepLR scheduler that reduces the learning rate by a factor of \(0.1\) at the 11th and 20th epochs. All experiments are conducted on two NVIDIA A100 (40GB) GPUs with a batch size of \(1\) per GPU (total batch size of \(2\)).

\section{Experimental}
\subsection{Setup}
\textbf{Datasets.} 
We conduct experiments on the SDGOD benchmark~\cite{wu2022single}, which contains five weather conditions: Daytime-Clear, Daytime-Foggy, Dusk-Rainy, Night-Clear, and Night-Rainy. Daytime-Clear is used as the source domain, with 19,395 images for training and 8,313 for testing. The remaining four conditions serve as unseen target domains, including 3,775 images in Daytime-Foggy, 3,501 in Dusk-Rainy, 26,158 in Night-Clear, and 2,494 in Night-Rainy. The dataset provides annotations for seven object categories: person, car, bike, rider, motor, bus, and truck.

\textbf{Evaluation metric.}
For evaluation, we adopt the mean Average Precision (mAP) as the primary performance metric. Following the standard evaluation protocol in single-domain generalized object detection, all mAP values are computed at an IoU threshold of 50\% (mAP@50), ensuring a fair and consistent comparison with previous methods.

\subsection{Comparison with Existing Methods}

We conduct a systematic comparison between the proposed method and a range of representative SDGOD approaches, with the results summarized in Table~\ref{tab:avg}. For clarity, the best results are highlighted in bold, while the second-best results among directly comparable methods are underlined. The compared methods mainly fall into three categories:
(1) methods built upon two-stage detection frameworks, including SDGOD~\cite{wu2022single}, CLIP the Gap~\cite{vidit2023clip}, SRCD~\cite{rao2024srcd}, DivAlign~\cite{danish2024improving}, G-NAS~\cite{wu2024g}, and UFR~\cite{liu2024unbiased}.
(2) general-purpose DETR-family detectors, including DINO~\cite{zhang2022dino}, Co-DETR~\cite{zong2022detrs}, Frozen-DETR~\cite{fu2024frozen}, and RT-DETR-v4~\cite{liao2025rt}, which distills semantic information from DINOv3;
(3) DETR extensions specifically designed for domain-generalized object detection, such as DG-DETR~\cite{hwang2025dg} and SA-DETR~\cite{han2025style}, both of which integrate style perturbation and explicit domain augmentation strategies to increase source-domain diversity and thereby improve adaptation to unseen domains.

Overall, our method achieves stable and significant performance gains under two different detection frameworks (DINO and Co-DETR), demonstrating strong generality. Specifically, compared with the baseline, it attains the largest average improvement under the DINO framework ($\uparrow$8.0\% mAP) and also delivers substantial gains under Co-DETR ($\uparrow$6.6\% mAP), achieving the best overall average performance (50.8\%). Relative to the second-best methods, it further improves performance by 5.2\% and 3.4\% under the DINO and Co-DETR frameworks, respectively. These results indicate that our method consistently enhances cross-domain detection robustness and generalization across diverse frameworks.

\begin{table}[t!]
\centering
\scriptsize
\setlength{\tabcolsep}{2pt}
\caption{Generalization Results on Five Different Domains.} \vspace{-7pt}
\begin{tabular}{l|ccccc|c}
\hline
Methods & D-Clear & N-Clear & D-Foggy & D-Rainy & N-Rainy & Avg. \\
\hline
SDGOD \cite{wu2022single}   &56.1&36.6&33.5&28.2&16.6&34.2\\
CLIP the Gap \cite{vidit2023clip}  &52.5&36.9&38.5&32.3&18.7&35.8\\
SRCD \cite{rao2024srcd}  &-&36.7&35.9&28.8&17.0&29.6\\
DivAlign~\cite{danish2024improving}   &52.8&42.5&37.2&38.1&24.1&38.9\\
G-NAS~\cite{wu2024g}   &58.4&45.0&36.4&35.1&17.4&38.5\\
UFR \cite{liu2024unbiased}  &58.6&40.8&39.6&33.2&19.2&38.3\\
\hline
DINO \cite{zhang2022dino}  &64.6&46.0&40.8&37.7&18.1&41.4\\
SA-DETR \cite{han2025style} (DINO)    &64.8&45.3&43.7&45.7&21.7&44.2\\
VFM$^{4}$SDG (DINO)    &\textbf{67.9}&\textbf{54.5}&\textbf{47.2}&\textbf{48.3}&\textbf{29.2}&\textbf{49.4}\\
\hline
DG-DETR \cite{hwang2025dg} &61.2&47.6&38.5&42.1&25.6&43.0\\
RT-DETR-v4 \cite{liao2025rt}  &65.9&52.2&43.5&47.7&27.5&47.4\\
\hline
Co-DETR \cite{zong2022detrs}  &68.1&50.4&43.4&40.8&18.3&44.2\\
Frozen-DETR \cite{fu2024frozen} (Co-DETR) &66.4&50.2&44.0&43.2&25.4&45.8\\
VFM$^{4}$SDG (Co-DETR)  &\textbf{71.0}&\textbf{56.1}&\textbf{47.4}&\textbf{50.7}&\textbf{29.0}&\textbf{50.8}\\
\hline
\end{tabular}\vspace{-10pt}
\label{tab:avg}
\end{table}

\textbf{Results on Daytime-Clear scene.}
Specifically, we evaluate the detection performance of all methods on the source domain. As shown in Table~\ref{tab:daytime_clear}, DETR-based approaches consistently outperform Faster R-CNN--based methods, indicating that transformer architectures offer superior modeling capacity and representation ability in this setting.
More importantly, VFM$^{4}$SDG built upon Co-DETR achieves the highest mAP of 71.0\% on the source domain, surpassing the original Co-DETR by 2.9\% and Frozen-DETR by 4.6\%, clearly demonstrating the effectiveness of the introduced visual priors in enhancing detection performance. Meanwhile, under the DINO detector, the proposed method also yields stable and substantial gains ($\uparrow$3.3\%) and further outperforms the DINO-based SA-DETR by 3.1\% mAP, highlighting its strong generality and consistent advantages across different DETR architectures.

\begin{table}[t!]
\centering
\scriptsize
\setlength{\tabcolsep}{2pt}
\caption{Quantitative results (\%) on the \textbf{Daytime-Clear} scene.}\vspace{-7pt}
\begin{tabular}{l|ccccccc|c}
\hline
Methods & Bus & Bike & Car & Motor & Person & Rider & Truck & mAP \\
\hline
SDGOD \cite{wu2022single}  & 68.8 & 50.9 & 53.9 & 56.2 & 41.8 & 52.4 & 68.7 & 56.1 \\
CLIP the Gap \cite{vidit2023clip} & 55.0 & 47.8 & 67.5 & 46.7 & 49.4 & 46.7 & 54.7 & 52.5 \\
DivAlign~\cite{danish2024improving}    &-&-&-&-&-&-&-&52.8\\
G-NAS~\cite{wu2024g}   &-&-&-&-&-&-&-&58.4\\
UFR \cite{liu2024unbiased}  & 66.8 & 51.0 & 70.6 & 55.8 & 49.8 & 48.5 & 67.4 & 58.6 \\
\hline
DINO \cite{zhang2022dino}  &64.3&55.4&85.7&54.9&69.5&55.7&66.5&64.6\\  
SA-DETR \cite{han2025style} (DINO)     &64.6&56.3&85.6&53.3&69.4&57.7&67.0&64.8  \\  
VFM$^{4}$SDG (DINO)    &\textbf{69.4}&\textbf{59.1}&\textbf{85.9}&\textbf{59.9}&\textbf{70.7}&\textbf{59.8}&\textbf{70.3}&\textbf{67.9} \\  
\hline
DG-DETR \cite{hwang2025dg} &62.5&50.7&83.5&51.9&61.6&53.3&64.9&61.2\\
RT-DETR-v4 \cite{liao2025rt}  &67.3 &55.3 &85.1 &61.2 &65.5 &59.0 &68.2 &65.9 \\
\hline
Co-DETR \cite{zong2022detrs}  &67.2&59.8&\textbf{87.4}&58.0&72.8&62.8&69.1&68.1 \\
Frozen-DETR \cite{fu2024frozen} (Co-DETR)  &65.7&58.9&85.9&57.8&69.9&59.1&67.7&66.4\\  
VFM$^{4}$SDG (Co-DETR)   &\textbf{71.3}&\textbf{63.1}&\textbf{87.4}&\textbf{65.0}&\textbf{73.5}&\textbf{64.3}&\textbf{72.4}&\textbf{71.0}\\
\hline
\end{tabular}\vspace{-10pt}
\label{tab:daytime_clear}
\end{table}


\textbf{Cross-domain detection results analysis.}
We systematically evaluate the cross-domain generalization performance of the VFM$^{4}$SDG on four unseen target domains (Night-Clear, Daytime-Foggy, Dusk-Rainy, and Night-Rainy), with results reported in Tables~\ref{tab:night_clear}, \ref{tab:daytime_foggy}, \ref{tab:dusk_rainy}, and \ref{tab:night_rainy}. Overall, VFM$^{4}$SDG achieves the best or near-best performance across all target domains and consistently delivers significant gains under both DINO- and Co-DETR-based detection frameworks, demonstrating the strong generality of the proposed dual-prior framework across different detectors and complex environments.

A closer examination of domain-specific results shows that VFM$^{4}$SDG consistently yields stable and substantial improvements over existing methods across diverse appearance variations. In the relatively mild Night-Clear scenario, the proposed method significantly outperforms the baseline in terms of overall mAP (DINO: $\uparrow$8.5\% / Co-DETR: $\uparrow$5.7\%). Under the Daytime-Foggy condition with pronounced appearance degradation, VFM$^{4}$SDG maintains clear advantages (DINO: $\uparrow$6.4\% / Co-DETR: $\uparrow$4.0\%). When facing the more challenging Dusk-Rainy scenario, which involves compounded illumination changes and rainfall, the gains become even more pronounced (DINO: $\uparrow$10.6\% / Co-DETR: $\uparrow$9.9\%). Notably, in the most severe and challenging Night-Rainy scenario, the proposed method achieves the largest performance improvements under both detection frameworks (DINO: $\uparrow$11.1\% / Co-DETR: $\uparrow$10.7\%), indicating strong robustness under complex domain shifts. Moreover, VFM$^{4}$SDG also consistently outperforms the second-best competing methods by a clear margin.

These results demonstrate that the proposed approach not only maintains stable cross-domain detection performance across diverse weather and imaging conditions, but also effectively mitigates omission-dominated performance degradation under severe domain shifts. This behavior aligns closely with our preceding analysis and further confirms that introducing cross-domain stable visual priors and jointly modeling representation learning and query alignment is critical for improving single-domain generalized object detection.

\begin{table}[t!]
\centering
\scriptsize
\setlength{\tabcolsep}{2pt}
\caption{Quantitative results (\%) on the \textbf{Night-Clear} scene.}\vspace{-7pt}
\begin{tabular}{l|ccccccc|c}
\hline
Methods & Bus & Bike & Car & Motor & Person & Rider & Truck & mAP \\
\hline
SDGOD \cite{wu2022single}  & 40.6 & 35.1 & 50.7 & 19.7 & 32.1 & 43.3 & 42.4 & 36.6 \\
CLIP the Gap \cite{vidit2023clip} & 37.7 & 34.3 & 58.0 & 17.2 & 37.6 & 28.5 & 42.9 & 36.9 \\
SRCD \cite{rao2024srcd}  &43.1&32.5&52.3&20.1&34.8&31.5&42.9&36.7\\
DivAlign~\cite{danish2024improving}   &-&-&-&-&-&-&-&42.5\\
G-NAS~\cite{wu2024g}   &46.9&40.5&67.5&26.5&50.7&35.4&47.8&45.0\\
UFR \cite{liu2024unbiased}   & 43.6 & 38.1 & 66.1 & 14.7 & 49.1 & 26.4 & 47.5 & 40.8 \\
\hline
DINO \cite{zhang2022dino} &47.1&43.4&73.4&19.9&57.4&31.5&49.5&46.0  \\  
SA-DETR \cite{han2025style} (DINO)  &48.9&42.7&72.9&14.3&56.8&32.5&49.0&45.3  \\  
VFM$^{4}$SDG (DINO)   &\textbf{59.6}&\textbf{51.0}&\textbf{74.1}&\textbf{32.8}&\textbf{61.4}&\textbf{40.9}&\textbf{61.7}&\textbf{54.5} \\
\hline
DG-DETR \cite{hwang2025dg} &51.2&44.7&73.4&23.0&51.9&36.1&52.5&47.6\\
RT-DETR-v4 \cite{liao2025rt}  &55.9 &49.7 &75.6 &28.6 &56.5 &42.5 &56.9 &52.2 \\
\hline
Co-DETR \cite{zong2022detrs}  &50.5&47.6&75.9&25.1&61.6&39.4&52.5&50.4\\
Frozen-DETR \cite{fu2024frozen} (Co-DETR)  &52.1&47.4&74.7&27.4&59.6&37.2&53.5&50.2\\  
VFM$^{4}$SDG (Co-DETR)   &\textbf{58.6}&\textbf{53.6}&\textbf{76.3}&\textbf{35.5}&\textbf{63.2}&\textbf{44.1}&\textbf{61.2}&\textbf{56.1}  \\
\hline
\end{tabular}\vspace{-10pt}
\label{tab:night_clear}
\end{table}

\begin{table}[t!]
\centering
\scriptsize
\setlength{\tabcolsep}{2pt}
\caption{Quantitative results (\%) on the \textbf{Daytime-Foggy} scene.}\vspace{-7pt}
\begin{tabular}{l|ccccccc|c}
\hline
Methods & Bus & Bike & Car & Motor & Person & Rider & Truck & mAP \\
\hline
SDGOD \cite{wu2022single}   & 32.9 & 28.0 & 48.8 & 29.8 & 32.5 & 35.2 & 24.1 & 33.5 \\
CLIP the Gap \cite{vidit2023clip}  & 36.2 & 34.2 & 57.9 & 34.0 & 38.7 & 43.8 & 25.1 & 38.5 \\
SRCD \cite{rao2024srcd}  &36.4&30.1&52.4&31.3&33.4&40.1&27.7&35.9\\
DivAlign~\cite{danish2024improving}   &-&-&-&-&-&-&-&37.2\\
G-NAS~\cite{wu2024g}   &32.4&31.2&57.7&31.9&38.6&38.5&24.5&36.4\\
UFR \cite{liu2024unbiased}    & 36.9 & 35.8 & 61.7 & 33.7 & 39.5 & 42.2 & 27.5 & 39.6 \\
\hline
DINO \cite{zhang2022dino}  &37.7&33.2&65.2&33.1&45.7&45.3&25.6&40.8\\  
SA-DETR \cite{han2025style} (DINO)    &40.5&33.8&\textbf{68.7}&35.7&\textbf{49.5}&48.5&29.0&43.7  \\  
VFM$^{4}$SDG (DINO)    &\textbf{50.9}&\textbf{34.6}&66.0&\textbf{41.6}&48.1&\textbf{50.4}&\textbf{38.9}&\textbf{47.2} \\
\hline
DG-DETR \cite{hwang2025dg} &36.4&28.9&64.4&31.2&40.3&39.9&28.8&38.5\\
RT-DETR-v4 \cite{liao2025rt}  &33.1 &\textbf{39.6} &\textbf{68.7} &39.1 &44.9 &44.9 &33.9 &43.5 \\
\hline
Co-DETR \cite{zong2022detrs}   &37.4&36.4&67.5&36.9&48.5&49.4&27.9&43.4\\
Frozen-DETR \cite{fu2024frozen} (Co-DETR)  &40.5&35.3&67.0&37.9&47.7&48.0&32.2&44.0\\  
VFM$^{4}$SDG (Co-DETR)    &\textbf{44.9}&37.4&68.5&\textbf{44.5}&\textbf{49.7}&\textbf{51.2}&\textbf{35.6}&\textbf{47.4}  \\
\hline
\end{tabular}\vspace{-10pt}
\label{tab:daytime_foggy}
\end{table}

\begin{table}[t!]
\centering
\scriptsize
\setlength{\tabcolsep}{2pt}
\caption{Quantitative results (\%) on the \textbf{Dusk-Rainy} scene.}\vspace{-7pt}
\begin{tabular}{l|ccccccc|c}
\hline
Methods & Bus & Bike & Car & Motor & Person & Rider & Truck & mAP \\
\hline
SDGOD \cite{wu2022single}  & 37.1 & 19.6 & 50.9 & 13.4 & 19.7 & 16.7 & 40.7 & 28.2 \\
CLIP the Gap \cite{vidit2023clip}  & 37.8 & 22.8 & 60.7 & 16.8 & 26.8 & 18.7 & 42.4 & 32.3 \\
SRCD \cite{rao2024srcd}   &39.5&21.4&50.6&11.9&20.1&17.6&40.5&28.8\\
DivAlign~\cite{danish2024improving}   &-&-&-&-&-&-&-&38.1\\
G-NAS~\cite{wu2024g}   &44.6&22.3&66.4&14.7&32.1&19.6&45.8&35.1\\
UFR \cite{liu2024unbiased}   & 37.1 & 21.8 & 67.9 & 16.4 & 27.4 & 17.9 & 43.9 & 33.2 \\ 
\hline
DINO \cite{zhang2022dino}  &42.2&29.0&75.1&17.1&37.9&15.8&46.6&37.7\\  
SA-DETR \cite{han2025style} (DINO)   &51.2&32.3&\textbf{77.0}&31.4&\textbf{49.3}&25.7&53.0&45.7\\  
VFM$^{4}$SDG (DINO)    &\textbf{56.3}&\textbf{34.7}&76.9&\textbf{34.5}&45.6&\textbf{29.2}&\textbf{60.9}&\textbf{48.3} \\
\hline
DG-DETR \cite{hwang2025dg}  &51.0&28.1&77.7&20.1&42.0&19.9&55.7&42.1\\
RT-DETR-v4 \cite{liao2025rt} &55.0 &29.3 &79.8 &\textbf{41.6} &44.3 &23.7 &60.3 &47.7 \\
\hline
Co-DETR \cite{zong2022detrs}  &48.4&28.2&77.6&21.0&38.7&21.6&50.6&40.8\\
Frozen-DETR \cite{fu2024frozen} (Co-DETR)  &49.8&30.9&77.2&24.9&41.7&23.4&54.6&43.2\\  
VFM$^{4}$SDG (Co-DETR)  &\textbf{57.0}&\textbf{40.0}&\textbf{79.0}&36.6&\textbf{47.3}&\textbf{33.6}&\textbf{61.1}&\textbf{50.7}\\
\hline
\end{tabular}\vspace{-10pt}
\label{tab:dusk_rainy}
\end{table}

\begin{table}[t!]
\centering
\scriptsize
\setlength{\tabcolsep}{2pt}
\caption{Quantitative results (\%) on the \textbf{Night-Rainy} scene.}\vspace{-7pt}
\begin{tabular}{l|ccccccc|c}
\hline
Methods & Bus & Bike & Car & Motor & Person & Rider & Truck & mAP \\
\hline
SDGOD \cite{wu2022single}   & 24.4 & 11.6 & 29.5 & 9.8 & 10.5 & 11.4 & 19.2 & 16.6 \\
CLIP the Gap \cite{vidit2023clip}  & 28.6 & 12.1 & 36.1 &  9.2 & 13.8 & 12.9 & 22.9 & 18.7 \\
SRCD \cite{rao2024srcd}   &26.5&12.9&32.4&0.8&10.2&12.5&24.0&17.0\\
DivAlign~\cite{danish2024improving}   &-&-&-&-&-&-&-&24.1\\
G-NAS~\cite{wu2024g}   &28.6&9.8&38.4&0.1&13.8&9.8&21.4&17.4\\
UFR \cite{liu2024unbiased}  & 29.9 & 11.8 & 36.1 &  9.4 & 13.1 & 10.5 & 23.3 & 19.2 \\
\hline
DINO \cite{zhang2022dino}  &29.7&13.1&43.0&0.9&12.6&5.8&21.6&18.1\\  
SA-DETR \cite{han2025style} (DINO)    &32.0&14.6&\textbf{45.8}&0.5&\textbf{23.2}&9.3&26.6&21.7\\  
VFM$^{4}$SDG (DINO)   &\textbf{50.9}&\textbf{16.0}&42.7&\textbf{11.3}&22.9&\textbf{18.3}&\textbf{41.9}&\textbf{29.2} \\
\hline
DG-DETR \cite{hwang2025dg}  &39.1&19.6&51.3&4.5&18.3&14.0&32.7&25.6\\
RT-DETR-v4 \cite{liao2025rt}  &44.3 &10.4 &\textbf{57.2} &4.8 &\textbf{23.7} &12.4 &39.5&27.5 \\
\hline
Co-DETR \cite{zong2022detrs}  &28.7&8.9&44.8&0.9&14.7&7.0&23.3&18.3\\
Frozen-DETR \cite{fu2024frozen} (Co-DETR)  &40.5&15.7&50.9&5.4&22.0&11.4&32.3&25.4\\  
VFM$^{4}$SDG (Co-DETR)     &\textbf{47.1}&\textbf{22.5}&47.6&\textbf{6.5}&22.6&\textbf{16.1}&\textbf{40.8}&\textbf{29.0}\\
\hline
\end{tabular}\vspace{-10pt}
\label{tab:night_rainy}
\end{table}

\textbf{Comparison with data augmentation methods.}
As shown in Table~\ref{tab:aug}, we select several recent representative domain augmentation methods, including ABA~\cite{cheng2023adversarial}, MAD~\cite{xu2023multi}, OA-DG~\cite{lee2024object}, PhysAug~\cite{xu2025physaug}, SRA~\cite{xiao2025sample}, and NP~\cite{fan2023towards}, and apply them to Co-DETR for comparison with our approach. These methods typically introduce perturbations in the image or feature space to synthesize training samples with diverse styles, thereby exposing the model to broader potential domain variations during training and enhancing its generalization capability to unseen target domains.

The results reveal two key observations. 
First, although various data augmentation strategies improve the Co-DETR's cross-domain performance to some extent, their overall gains remain substantially lower than those of VFM$^{4}$SDG. In average mAP, the strongest augmentation method, NP, raises performance from 44.2\% to 47.0\%, whereas VFM$^{4}$SDG further increases it to 50.8\%, maintaining a clear margin of 3.8\% over NP.
Second, this performance gap becomes even more pronounced in more challenging target domains. In the most severe Night-Rainy scenario, where appearance variations are the most drastic, augmentation-based methods yield only limited improvements (NP: 23.1\%), while VFM$^{4}$SDG reaches 29.0\% ($\uparrow$ 5.9\%), demonstrating markedly stronger robustness. These findings indicate that merely expanding stylistic diversity during training is insufficient to fundamentally mitigate generalization degradation under severe domain shifts. In contrast, VFM$^{4}$SDG introduces cross-domain stable visual priors and jointly enhances relational modeling and query alignment stability, thereby delivering more consistent and substantial performance gains under complex domain shifts.

\begin{table}[t!]
\centering
\scriptsize
\setlength{\tabcolsep}{2pt}
\caption{Comparison with Data Augmentation Methods.}\vspace{-7pt}
\begin{tabular}{l|ccccc|c}
\hline
Methods & D-Clear & N-Clear & D-Foggy & D-Rainy & N-Rainy & Avg. \\
\hline
Co-DETR  &68.1&50.4&43.4&40.8&18.3&44.2  \\
~+ ABA \cite{cheng2023adversarial}     &68.4&52.0&44.3&46.5&21.4&46.5\\
~+ NP~\cite{fan2023towards}  &68.3&51.2&44.6&47.7&23.1&47.0\\
~+ MAD \cite{xu2023multi}       &67.9&50.8&45.2&41.6&19.0&44.9\\
~+ OA-DG \cite{lee2024object}      &66.7&47.9&42.1&40.7&17.8&43.0\\
~+ SRA \cite{xiao2025sample}       &67.3&49.2&44.7&41.1&21.3&44.7\\
~+ PhysAug \cite{xu2025physaug}   &67.4&48.7&44.4&41.1&21.6&44.6\\
VFM$^{4}$SDG (Co-DETR)  &\textbf{71.0}&\textbf{56.1}&\textbf{47.4}&\textbf{50.7}&\textbf{29.0}&\textbf{50.8}\\
\hline
\end{tabular}\vspace{-10pt}
\label{tab:aug}
\end{table}

\subsection{Ablation Study}
We systematically conduct ablation studies across all scenarios to comprehensively analyze the role and contribution of each component in VFM$^{4}$SDG.

\textbf{Effectiveness of individual components.}
We conduct a detailed component-wise effectiveness analysis of the proposed method under both the \textbf{Co-DETR}. Taking Table~\ref{tab:components} as an example, the first row corresponds to the baseline model; the second row augments the baseline with \textbf{stable relational distillation}; the third, fourth, and fifth rows respectively incorporate \textbf{category-prototype-guided cross-attention}, \textbf{DINOv3-guided cross-attention}, and their combination; the sixth row reports the performance when all proposed components are jointly applied.

Overall, each proposed component improves the cross-domain performance of the baseline to varying degrees, and clear complementarity emerges among them.
Introducing CSRPD increases the average mAP from 44.2\% to 45.8\%, yielding consistent gains across all target domains, particularly in the Night-Rainy scenario (from 18.3\% to 21.2\%). This demonstrates that distilling the cross-domain stable relational encoded in DINOv3 effectively alleviates representation instability under domain shift.
At the query modeling stage, incorporating either SIGA or CSGA individually also significantly outperforms the baseline. SIGA reinforces semantic identity constraints for queries, strengthening category-level guidance, whereas CSGA enhances spatial localization by introducing stable contextual modeling. Notably, CSGA brings more pronounced improvements in challenging scenarios such as Night-Clear, Dusk-Rainy, and Night-Rainy, highlighting the critical role of spatial context stability under severe domain shifts.
When combined as SCPQE, the average performance further rises to 50.4\%, revealing strong synergy between semantic guidance and spatial modeling and validating the design principle of `semantic identity first, spatial alignment next'.
Finally, jointly integrating CSRPD and SCPQE yields the best performance (50.8\% mAP) and achieves the highest results across all target domains. These findings confirm that relational stability in the encoder and query alignment stability in the decoder are complementary and jointly essential, and their interaction is key to enhancing detection robustness under severe domain shifts.

\begin{table}[t!]
\centering
\scriptsize
\setlength{\tabcolsep}{2pt}
\caption{Ablation study of different components.}\vspace{-7pt}
\begin{tabular}{l|ccccc|c}
\hline
Methods & D-Clear & N-Clear & D-Foggy & D-Rainy & N-Rainy & Avg. \\
\hline
Co-DETR  &68.1&50.4&43.4&40.8&18.3&44.2\\
~+ CSRPD       &68.8&51.1&45.8&42.0&21.2&45.8\\
~+ SIGA    &70.5&52.2&44.6&42.5&20.3&46.0\\
~+ CSGA    &70.1&55.2&46.5&47.2&26.9&49.2\\
~+ SCPQE (SIGA + CSGA)      &70.8&55.6&46.8&50.1&28.6&50.4\\
VFM$^{4}$SDG (Co-DETR)      &\textbf{71.0}&\textbf{56.1}&\textbf{47.4}&\textbf{50.7}&\textbf{29.0}&\textbf{50.8}\\
\hline
\end{tabular}\vspace{-15pt}
\label{tab:components}
\end{table}

\textbf{Relational Distillation vs. Semantic Distillation.}  
As shown in Table~\ref{tab:KD}, we further analyze different DINOv3-based transfer strategies using Co-DETR as the baseline, including semantic distillation (KD-Sema) and relational distillation (KD-Rela). Here, `to Backbone' and `to Encoder' indicate that the distillation constraint is applied to backbone output features and encoder outputs, respectively.
Overall, semantic distillation (KD-Sema) significantly improves the baseline. When applied to the backbone, it raises the average mAP from 44.2\% to 49.4\%; when applied to the encoder, it achieves 49.0\%. This shows that the high-quality semantic representations provided by DINOv3 effectively enhance the detector's representational capacity.

In contrast, relational distillation (KD-Rela) yields larger gains when applied to the encoder, achieving the best overall performance of 50.8\% mAP and outperforming all other variants across target domains. Its advantage is especially evident under more severe domain shifts, such as Dusk-Rainy and Night-Rainy. This suggests that preserving cross-domain stable object-background and inter-instance relations is more critical for generalization than directly aligning absolute semantic representations.
A closer analysis shows that semantic distillation strengthens semantic encoding by aligning absolute feature distributions. However, because it relies solely on single-source data (Daytime-Clear), it may inadvertently reinforce source-domain bias and thus limit generalization to unseen domains. In contrast, relational distillation focuses on the consistency of object-background and inter-instance relations rather than domain-dependent semantic magnitudes, leading to stronger cross-domain stability.

In summary, these results further support our central claim: under the single-domain generalization setting, distilling cross-domain stable relations is more effective than directly transferring absolute semantic information for mitigating representation instability caused by domain shift, thereby yielding superior cross-domain robustness.





\begin{table}[t!]
\centering
\scriptsize
\setlength{\tabcolsep}{2pt}
\caption{Comparison of Different DINOv3 Transfer Strategies.}\vspace{-7pt}
\begin{tabular}{l|ccccc|c}
\hline
VFMs & D-Clear & N-Clear & D-Foggy & D-Rainy & N-Rainy & Avg. \\
\hline
Co-DETR   &68.1&50.4&43.4&40.8&18.3&44.2  \\
~by KD-Sema (to Backbone)   &70.1&55.2&46.2&48.3&27.1&49.4 \\
~by KD-Sema (to Encoder)  &69.9&55.0&46.5&47.5&26.0&49.0\\
~\textbf{by KD-Rela (to Encoder)}   &\textbf{71.0}&\textbf{56.1}&\textbf{47.4}&\textbf{50.7}&\textbf{29.0}&\textbf{50.8}\\
\hline
\end{tabular}\vspace{-10pt}
\label{tab:KD}
\end{table}

\textbf{Sensitivity analysis of the relational distillation loss weight.}
As shown in Table~\ref{tab:weights}, we conduct a sensitivity analysis on the relational distillation loss weight $\lambda$. Specifically, $\lambda$ is set to $0.1$, $0.3$, $0.5$, $0.8$, $1.0$, $1.2$, and $1.5$. The results show that the model remains stable over a relatively wide range of values. Among them, $\lambda=1.0$ achieves the best average mAP (50.8\%), while $\lambda=0.8$ and $0.5$ also yield comparable results, indicating that the proposed method is robust to this hyperparameter.
This suggests that relational distillation should be introduced with a moderate weight. When the weight is too small, the constraint on cross-domain stable relations is insufficient, making it difficult to fully exploit the relational priors. In contrast, an excessively large weight may suppress the detector's learning of task-relevant discriminative features and thus limit overall performance. Therefore, achieving a proper balance between relational constraint and detection objective optimization is crucial. Under our setting, $\lambda=1.0$ provides the best trade-off between the two.


\begin{table}[t!]
\centering
\scriptsize
\setlength{\tabcolsep}{4.5pt}
\caption{Effect of the weight of loss $\lambda$ in CRPD.}\vspace{-7pt}
\begin{tabular}{l|ccccc|c}
\hline
$\lambda$ & D-Clear & N-Clear & D-Foggy & D-Rainy & N-Rainy & Avg. \\
\hline
0.1     &70.4&55.5&47.0&50.3&25.3&49.7\\
0.3     &70.8&55.3&47.0&50.6&27.1&50.2\\
0.5     &70.6&55.9&46.9&50.4&\textbf{29.1}&50.6\\
0.8     &70.8&55.8&\textbf{47.5}&\textbf{50.8}&28.8&50.7\\
1.0     &\textbf{71.0}&\textbf{56.1}&47.4&50.7&29.0&\textbf{50.8}\\
1.2     &70.7&55.8&47.2&50.6&28.3&50.5\\
1.5     &70.6&55.5&46.8&50.6&28.0&50.3\\
\hline
\end{tabular}\vspace{-10pt}
\label{tab:weights}
\end{table}

\textbf{Sensitivity analysis of relational distillation across feature layers.}
As shown in Table~\ref{tab:layers}, we systematically analyze the effect of the number of feature levels involved in relational distillation under the Co-DETR framework. After reconstruction, the encoder outputs five multi-scale feature levels aligned with the backbone in both resolution and hierarchy, denoted as $\{0,1,2,3,4\}$. Among them, $\{4\}$ corresponds to the highest semantic level and typically contains the richest semantic information. We therefore take $\{4\}$ as the base distillation level and progressively extend to lower levels, evaluating $\{4\}$, $\{4,3\}$, $\{4,3,2\}$, $\{4,3,2,1\}$, and $\{4,3,2,1,0\}$.
Overall, performance improves steadily as more levels participate in relational distillation, reaching the best result when all five levels are jointly distilled. This suggests that constraining only high-level semantic features provides guidance but is insufficient to preserve multi-scale relational information under domain shift. In contrast, applying relational distillation across all scales establishes consistent relational constraints across semantic hierarchies, enabling more complete transfer of the cross-domain stable relational priors encoded in DINOv3.

Moreover, since Co-DETR interacts with multi-scale tokens during decoding, aligning relations at only part of the scales may lead to relational geometric inconsistency across levels. Full-level relational distillation instead enforces cross-level consistent relational supervision in the multi-scale feature space, providing a more stable relational foundation for the subsequent SCPQE guided by category prototypes and DINOv3 features, and thereby further improving detection performance and cross-domain generalization.


\begin{table}[t!]
\centering
\scriptsize
\setlength{\tabcolsep}{4.5pt}
\caption{Effect of the number of feature layers distilled in CRPD.}\vspace{-7pt}
\begin{tabular}{l|ccccc|c}
\hline
Layers & D-Clear & N-Clear & D-Foggy & D-Rainy & N-Rainy & Avg. \\
\hline
Co-DETR  &68.1&50.4&43.4&40.8&18.3&44.2  \\
\(\{4\}\)     &68.1&50.6&44.7&41.2&20.2&45.0  \\
\(\{4,3\}\)     &68.3&50.4&45.2&41.4&20.6&45.2 \\
\(\{4,3,2\}\)     &68.4&50.7&45.5&41.8&20.4&45.4 \\
\(\{4,3,2,1\}\)     &68.7&\textbf{51.1}&45.6&41.7&21.1&45.6 \\
\(\{4,3,2,1,0\}\)    &\textbf{68.8}&\textbf{51.1}&\textbf{45.8}&\textbf{42.0}&\textbf{21.2}&\textbf{45.8} \\
\hline
\end{tabular}\vspace{-15pt}
\label{tab:layers}
\end{table}

\textbf{Effect of using different VFMs.}
As shown in Table~\ref{tab:VFMs}, we further evaluate the generalization of the proposed method with different visual foundation models (VFMs) under the Co-DETR framework. In addition to DINOv3, we consider two representative models, BEIT-3 (Large)~\cite{wang2023image} and SAM3~\cite{carion2025sam}. These models differ substantially in training paradigm, data scale, and representational focus: BEIT-3 adopts multi-modal joint pretraining, emphasizing semantic alignment and cross-modal consistency; SAM3 is trained on large-scale segmentation data and, while also modeling cross-modal consistency, places greater emphasis on region-level structure and boundary modeling; DINOv3, by contrast, is pretrained on large-scale visual data in a self-supervised manner, highlighting cross-domain relational consistency and global semantic structure.
Experimental results show that, despite these differences in pretraining objectives and representation properties, the proposed method consistently improves performance across all VFMs. This indicates that the gains do not rely on the incidental advantage of any specific foundation model, but rather on effectively extracting cross-domain stable priors that are broadly present in VFMs and transferring them to the detector to enhance single-domain generalization.
Further analysis shows that SAM3 and DINOv3 yield more pronounced gains, whereas BEIT-3 provides relatively smaller improvements, which may be related to differences in pretraining scale and structural modeling capacity. Compared with DINOv3 and SAM3, which are trained on larger-scale visual data, BEIT-3 may be relatively limited in its ability to encode cross-domain relational structures.

In addition, to rule out the possibility that the performance gain simply comes from increased model size, we replace SCPQE with two self-attention modules (+2*SA) for comparison. Although the two variants differ slightly in implementation details, they have comparable learnable parameter counts overall. The results show that simply stacking attention layers does not bring similar gains, suggesting that the improvement comes from the proposed cross-domain stable visual prior guidance itself rather than parameter expansion.
Overall, these results verify the robustness and scalability of the proposed dual-prior framework. Rather than being a model-specific trick, it serves as a general prior-learning mechanism that complements diverse VFMs and improves single-domain generalization for object detection.

\begin{table}[t!]
\centering
\scriptsize
\setlength{\tabcolsep}{2pt}
\caption{Comparison of Different VFMs.}\vspace{-7pt}
\begin{tabular}{l|ccccc|c}
\hline
VFMs & D-Clear & N-Clear & D-Foggy & D-Rainy & N-Rainy & Avg. \\
\hline
Co-DETR  &68.1&50.4&43.4&40.8&18.3&44.2  \\
\textbf{Co-DETR + 2*SA} &68.3&50.2&43.4&42.0&17.4&44.3\\ 
\textbf{VFM$^{4}$SDG (BEIT-3)}  &68.7&50.6&43.9&41.8&21.1&45.2 \\ 
\textbf{VFM$^{4}$SDG (SAM3)}  &\textbf{71.2}&55.8&46.7&50.4&\textbf{29.1}&50.6 \\
\textbf{VFM$^{4}$SDG (DINOv3)}  &71.0&\textbf{56.1}&\textbf{47.4}&\textbf{50.7}&29.0&\textbf{50.8}\\
\hline
\end{tabular}\vspace{-10pt}
\label{tab:VFMs}
\end{table}

\subsection{Model Complexity and Inference Efficiency}  
We further analyze different methods from the perspectives of model complexity and inference efficiency. Under the same experimental setting, Co-DETR (baseline), Frozen-DETR, and VFM$^{4}$SDG are evaluated on an NVIDIA A100 GPU, with results reported in Table~\ref{tab:efficiency}.
Co-DETR has the fewest parameters and the fastest inference speed, reflecting its efficiency-oriented design. Frozen-DETR, by introducing a large frozen CLIP model, substantially increases the total parameter count and correspondingly reduces inference speed.
VFM$^{4}$SDG introduces prior-guided modules, leading to higher total and learnable parameter counts, while keeping the overall increase moderate. Although inference becomes slightly slower, the proposed method achieves clear gains in cross-domain detection performance, yielding a more practical balance between accuracy and efficiency.
More importantly, these gains do not stem from a simple increase in model size. Compared with Frozen-DETR, VFM$^{4}$SDG adds only about 3.7M learnable parameters yet achieves markedly better generalization, indicating that the improvement mainly comes from effective modeling and utilization of stable visual priors rather than parameter scaling.


\begin{table}[t!]
\centering
\caption{Model complexity and inference efficiency comparison.}\vspace{-7pt}
\label{tab:efficiency}
\begin{tabular}{l|ccc}
\hline
Method & Total Params (M) & Learnable Params (M) & FPS \\
\hline
Co-DETR & 64.51 & 64.24 & 10.49 \\
Frozen-DETR & 370.04 & 66.85 & 5.82 \\
VFM$^{4}$SDG & 373.98 & 70.56 & 4.05 \\
\hline
\end{tabular}\vspace{-15pt}
\end{table}

\begin{figure*}[t!]
	\centering
	\includegraphics[width=1.0\linewidth]{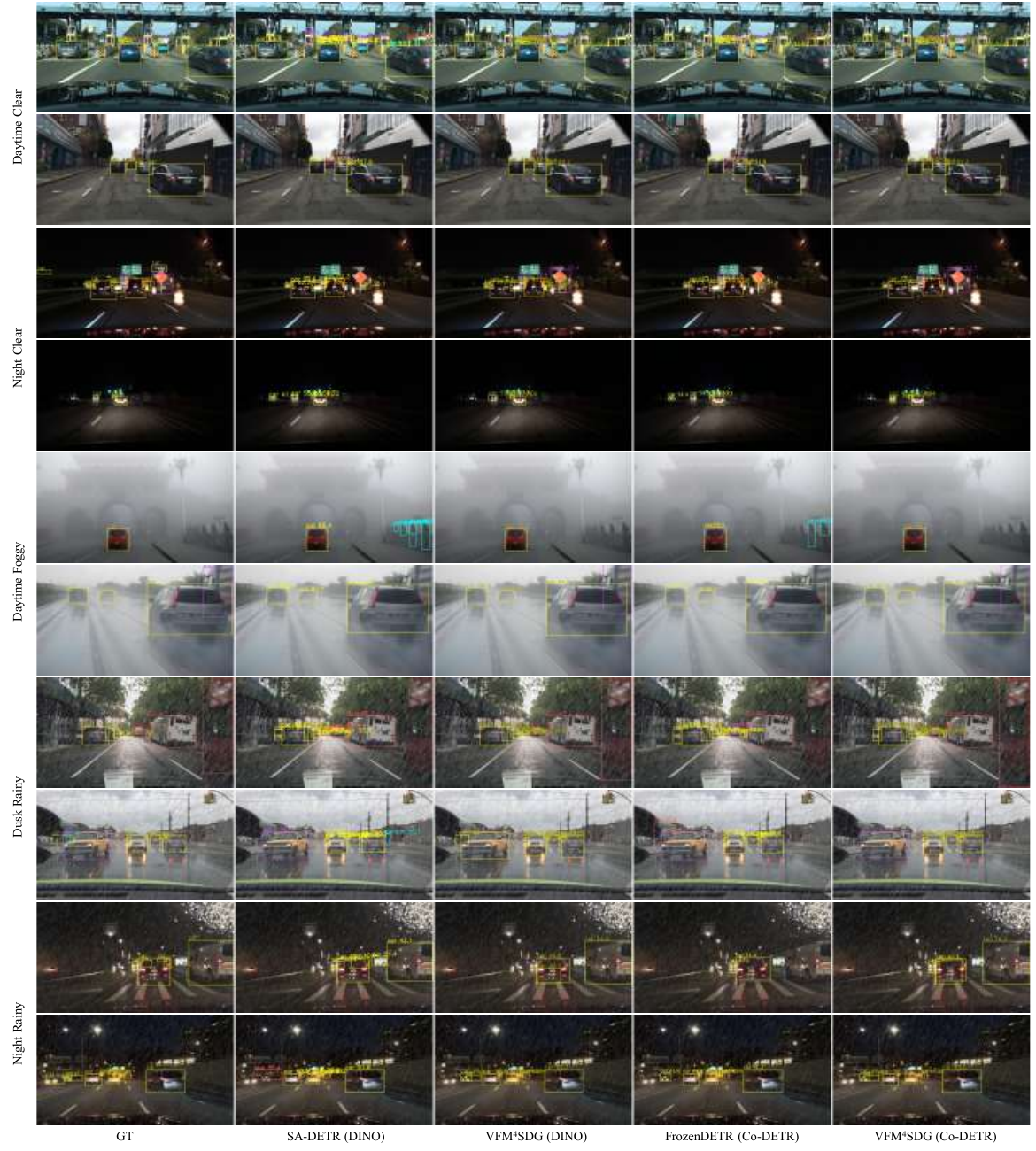}\vspace{-10pt}
	\caption{Qualitative comparisons under diverse domain conditions. We compare our detection results with state-of-the-art methods, with different categories indicated by distinct colors. From left to right: Ground Truth (GT), SA-DETR (DINO-based), VFM$^{4}$SDG (DINO-based), Frozen-DETR (Co-DETR-based), and VFM$^{4}$SDG (Co-DETR-based). The results span multiple domains, including Daytime-Clear, Daytime-Foggy, Dusk-Rainy, Night-Rainy, and Night-Clear. Compared with existing approaches, VFM$^{4}$SDG produces more complete detections under severe appearance variations, substantially reducing both missed detections and false positives. Overall predictions remain more stable and consistent across domains, demonstrating stronger cross-domain robustness.}
	\label{fig:vision}
	\vspace{-15pt}
\end{figure*}

\begin{figure*}[t!]
	\centering
	\includegraphics[width=0.95\linewidth]{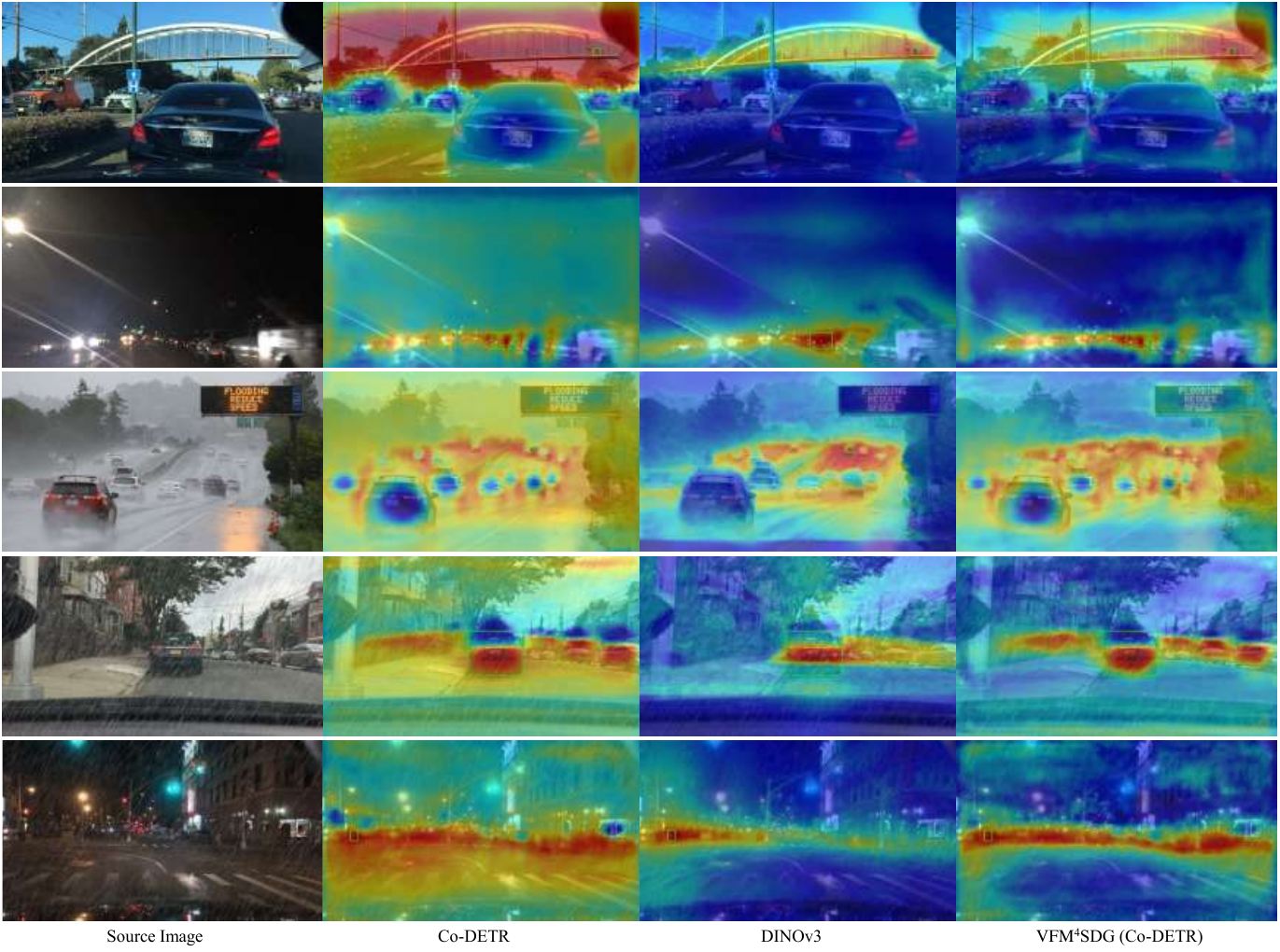}\vspace{-7pt}
	\caption{Visualization of Encoder Feature Responses under Domain Shift. From left to right: source image, Co-DETR encoder features, DINOv3 features, and VFM$^{4}$SDG (Co-DETR-based) encoder features. The visualizations correspond to the second encoder layer (relatively high spatial resolution). Warmer colors indicate stronger feature responses.}
	\label{fig:KD}
	\vspace{-10pt}
\end{figure*}

\begin{figure}[t!]
	\centering
	\includegraphics[width=0.9\linewidth]{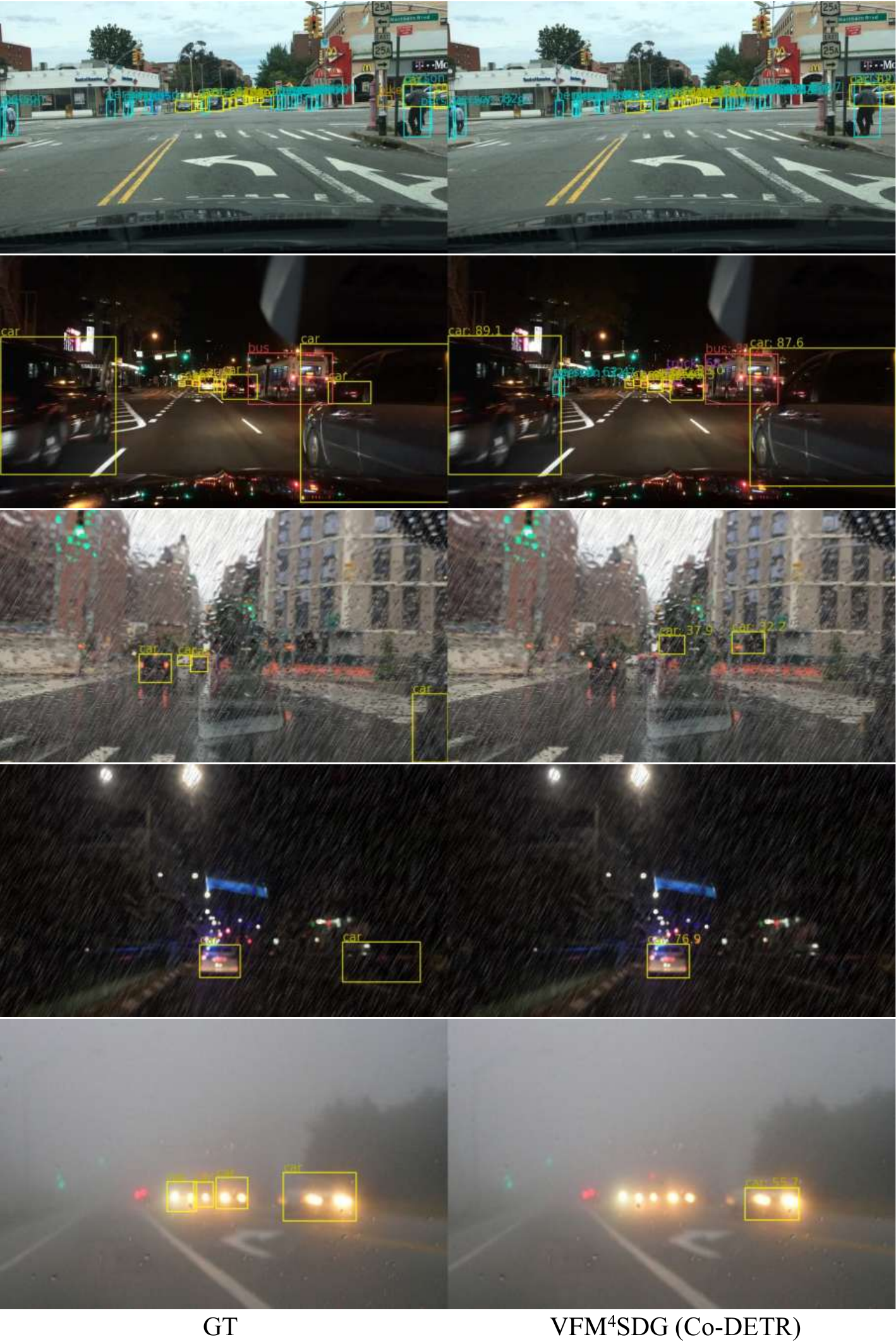}\vspace{-10pt}
	\caption{Representative failure cases of VFM$^{4}$SDG (Co-DETR-based) under challenging domain conditions. From left to right: GT and VFM$^{4}$SDG predictions. The examples include scenarios with heavy rain, nighttime illumination, dense fog, motion blur, small-scale objects, and severe occlusion. While the proposed method maintains robust detection performance in most cases, certain missed detections still occur under extreme visibility degradation or when objects are extremely small and partially occluded.}  
	\label{fig:failcase}
	\vspace{-15pt}
\end{figure}

\subsection{Visualization and Analysis}
As shown in Fig.~\ref{fig:vision}, the qualitative results further corroborate the quantitative findings. In relatively mild scenarios such as Daytime-Clear, all methods perform similarly overall, but VFM$^{4}$SDG detects small and distant objects more completely and yields more stable bounding box localization. As domain shift becomes more severe, competing methods exhibit clear performance degradation. Both SA-DETR and Frozen-DETR suffer from more frequent missed detections and unstable localization under low contrast, complex illumination, or partial occlusion, and this trend becomes even more pronounced in challenging scenes such as Night-Rainy.
In contrast, VFM$^{4}$SDG more consistently preserves valid object instances and maintains accurate localization across domains. This further suggests that stable relational modeling and instance-query alignment are crucial for cross-domain robustness under domain shift. By introducing cross-domain stable priors, the proposed dual-prior framework effectively alleviates representation instability caused by severe appearance changes, leading to more reliable cross-domain detection.

\textbf{Visualization of encoder feature responses under domain shift.}
As shown in Fig.~\ref{fig:KD}, these visualizations provide intuitive evidence for the role of relational distillation at the encoding stage. In the baseline Co-DETR, feature responses tend to spread into background regions under complex backgrounds or severe domain shifts, leading to insufficient foreground-background separation and indicating unstable instance representations across domains.
In contrast, DINOv3 exhibits clearer cross-domain relational consistency and stronger object-background discrimination, reflecting the cross-domain stable relations learned through large-scale pretraining. After introducing relational distillation, the encoder features of VFM$^{4}$SDG become more concentrated on semantic object regions while suppressing irrelevant background activations, resulting in a more compact and clearer overall structure.
Notably, the distilled features do not simply replicate those of DINOv3; instead, they incorporate cross-domain stable relational constraints while preserving the discriminative capacity required for detection. This suggests that relational distillation mainly transfers structural priors rather than absolute semantic values, thereby improving the stability of encoded representations under severe domain shifts.
Overall, these visualizations further support our central claim that stabilizing object-background and inter-instance relations at the encoding stage enhances foreground-background decoupling and thus improves cross-domain detection robustness.

\textbf{Failure Cases of VFM$^{4}$SDG under Severe Appearance Degradation.}
As shown in Fig.~\ref{fig:failcase}, these failure cases reflect the inherent challenges of cross-domain detection under extreme visual degradation. Although VFM$^{4}$SDG improves robustness by introducing stable cross-domain visual priors, reliable detection remains difficult when object contours and structural cues are severely degraded, such as in heavy rain, dense fog, or low-contrast nighttime conditions.
Most missed detections occur on small, heavily occluded, or distant objects, whose weak signals are more easily overwhelmed by environmental noise.
Overall, these cases highlight the difficulty of cross-domain detection under severe appearance degradation and suggest that, even with visual priors, there is still room for improvement under strong noise, heavy occlusion, or extremely small-object scenarios.

\section{Conclusion}
Rather than following the conventional representation-invariance perspective, this paper revisits SDGOD from the viewpoint of detector mechanisms. We argue that degradation under domain shift is not merely caused by appearance variation or insufficient category discrimination, but more fundamentally by reduced cross-domain detector stability: disrupted encoder-side relational structures further weaken decoder-side query-object binding.
Motivated by the finding that degradation is dominated by increasing missed detections, we propose VFM$^{4}$SDG, a dual-prior framework that introduces a frozen VFM as a degradation-driven cross-domain stability prior. Through Cross-domain Stable Relational Prior Distillation and Semantic-Contextual Prior-based Query Enhancement, VFM$^{4}$SDG compensates for encoder-side relational degradation and enhances decoder-side query-object binding, improving robustness in unseen domains.
Extensive experiments show that VFM$^{4}$SDG consistently outperforms existing methods under diverse adverse conditions and across different DETR-based architectures. Our study highlights the importance of jointly preserving relational stability and query-object binding stability for robust SDGOD, and offers new insight into using VFMs for robust visual perception.

\section{Limitations and Future Directions}  
Although substantial progress is achieved in generalization performance, existing SDGOD methods—including ours—generally incur considerable computational overhead, which limits their applicability in real-time scenarios. Future research therefore focuses on developing lightweight and real-time–oriented single-domain generalization frameworks that reconcile efficiency and robustness without introducing additional complexity at inference time.



\bibliographystyle{IEEEtran}
\bibliography{8-Reference}

\end{document}